%% file: main.tex
\ifdefined\pdfobjcompresslevel
\fi
\ifdefined\pdfcompresslevel
\fi
\ifdefined\pdfvariable
  \pdfvariable objcompresslevel=0
  \pdfvariable compresslevel=0
\fi

\documentclass[11pt]{article}

\usepackage[preprint]{acl}
\input{math_commands.tex}

\usepackage{times}
\usepackage{latexsym}

\usepackage[T1]{fontenc}

\usepackage[utf8]{inputenc}

\usepackage{microtype}

\usepackage{inconsolata}

\usepackage{caption}
\usepackage{subcaption}
\usepackage{enumitem}
\usepackage{flushend}
\usepackage{balance}
\usepackage{lineno}
\usepackage{url}
\usepackage{amsmath,amssymb,amsfonts}
\usepackage{algorithm}
\usepackage{algorithmic}
\usepackage{graphicx}
\usepackage{bbding}
\usepackage{textcomp}
\usepackage{xcolor}
\usepackage{hyperref}
\usepackage{colortbl}
\usepackage{amsthm}
\usepackage{float}
\usepackage{multirow}
\usepackage{multicol}
\usepackage{color}
\usepackage{bm}
\usepackage{bbm}
\usepackage[most]{tcolorbox}
\definecolor{lightgraybox}{gray}{0.95}

\usepackage{pifont}
\newcommand{\algcomment}[1]{\hfill $\triangleright$~#1}

\usepackage{array}
\newcolumntype{L}[1]{>{\raggedright\let\newline\\\arraybackslash\hspace{0pt}}m{#1}}
\newcolumntype{C}[1]{>{\centering\let\newline  \\\arraybackslash\hspace{0pt}}m{#1}}
\newcolumntype{R}[1]{>{\raggedleft\let\newline \\\arraybackslash\hspace{0pt}}m{#1}}

\usepackage[utf8]{inputenc} 
\usepackage[T1]{fontenc}    
\usepackage{hyperref}       
\usepackage{url}            
\usepackage{booktabs}       
\usepackage{amsfonts}       
\usepackage{nicefrac}       
\usepackage{microtype}      
\usepackage{xcolor}         
\usepackage{comment}
\usepackage{arydshln}

\ifodd 1
\newcommand{\czh}[1]{{\color{cyan}(czh: #1)}}
\else
\newcommand{\czh}[1]{}
\fi

\title{The Past Is Prologue: A Plug-in Controller for Selective Updates in Sequentially Evolving LLM Memory}

\author{
  Zihan Chen\textsuperscript{\ding{171}},  
  Songwei Dong\textsuperscript{\ding{171}},  
  Chengshuai Shi\textsuperscript{\ding{168}},  
  Peng Wang\textsuperscript{\ding{171}},  \\
  \textbf{Song Wang}\textsuperscript{\ding{169}},  
  \textbf{Cong Shen}\textsuperscript{\ding{171}},  
  \textbf{Jundong Li}\textsuperscript{\ding{171}} \\
  \textsuperscript{\ding{171}}University of Virginia,
  \textsuperscript{\ding{168}}Princeton University,
  \textsuperscript{\ding{169}}University of Central Florida,\\
  {\tt \{brf3rx, hxt5ap, pw7nc, cong, jundong\}@virginia.edu} \\ {\tt 
cs1083@princeton.edu} ,\quad 
  {\tt song.wang@ucf.edu}
}

\begin{document}
\maketitle
\begin{abstract}
Sequentially evolving LLM memory enables agents to reuse past experience, but existing systems usually deploy each locally generated memory update without checking whether it improves future behavior. As a result, updates that help the current task may overwrite useful knowledge, introduce over-specific rules, or bias the final memory toward recent examples. We propose Janus, a plug-in memory controller that decides whether to accept a candidate memory update or retain the previous memory.  To make this decision efficient, Janus uses a Memory Momentum Trigger to identify suspicious deviations in the memory-update trajectory, and compares old and new memories on a compact hybrid evaluation set of coverage, boundary, and fresh tasks instead of replaying the full history. Janus is method-agnostic and wraps existing updaters without changing their update rules. Across six datasets, two backbone LLMs, and two memory updaters, Janus improves average accuracy by +2.7 to +4.6 points over the corresponding base updaters. 
\end{abstract}

\section{Introduction}


Large language models (LLMs) are increasingly deployed as sequential task-solving agents that learn from past interactions through external memory~\citep{xiang2026systematic,fang2025memp,wei2025evo}. 
In this setting, the LLM repeatedly encounters tasks, produces answers, receives feedback, and updates its memory for future use. 
The resulting trajectories contain successful solutions, failed attempts, feedback signals, and intermediate reasoning traces, which can serve as reusable experience for improving future decisions~\citep{zhao2024expel,wei2025evo,suzgun2026dynamic,zhou2025memento}. 
This form of \emph{sequentially evolving memory} goes beyond static conversational recall: memory is not merely a record of past interactions, but a test-time adaptation mechanism that shapes the LLM's behavior on future tasks. 
Such systems are important for reasoning assistants~\citep{ho2025arcmemo}, tool-use agents~\citep{wang2025reinforcement}, and interactive decision-making systems~\citep{zheng2025skillweaver,agrawal2025gepa}, where performance depends not only on the current input but also on the accumulated experience.

\begin{figure}[t]
    \centering
    \begin{subfigure}{0.5\textwidth}
        \centering
        \includegraphics[width=\textwidth]{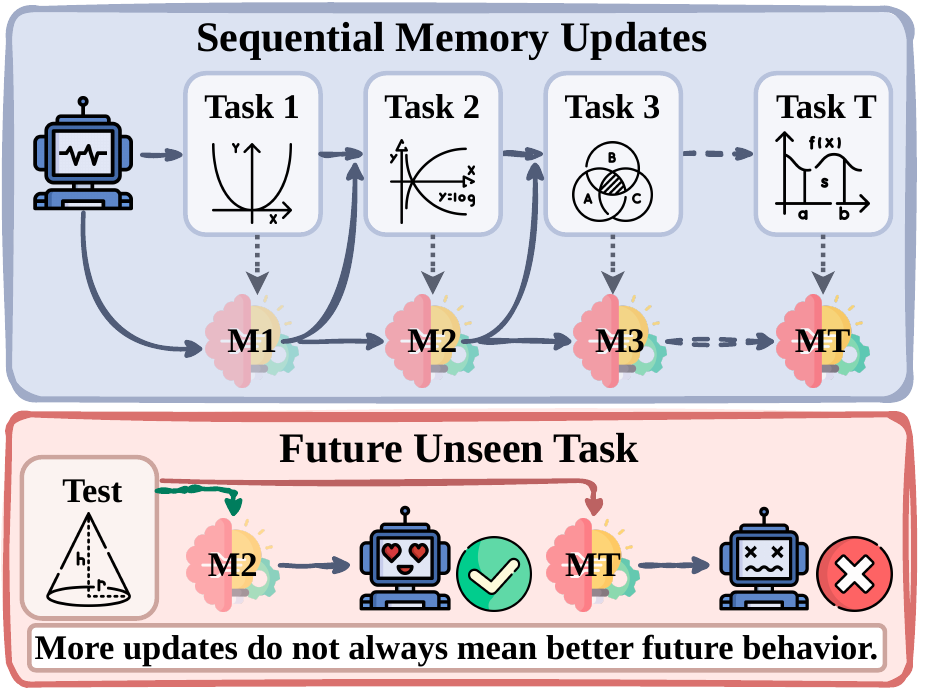}
    \end{subfigure}


    \begin{subfigure}{0.5\textwidth}
        \centering
        \includegraphics[width=\textwidth]{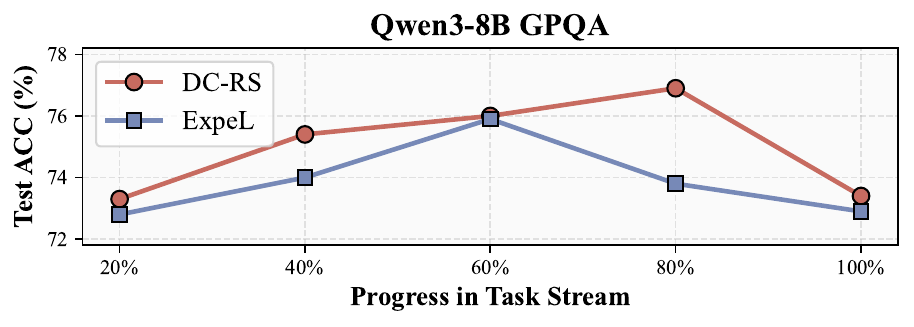}
    \end{subfigure}\vspace{-0.1in}
    \caption{\textbf{Top:} Existing sequential memory updates do not guarantee that the final memory will better support future tasks.
    \textbf{Bottom:} On GPQA, intermediate memory snapshots from two sequential memory methods exhibit non-monotonic test set performance, motivating the need to control which memory updates are deployed.
    }\vspace{-0.2in}
     \label{fig:teaser}
     
\end{figure}

A common mechanism behind these systems is to update memory using feedback from the current task~\cite{wang2025agent,wei2025evo}. 
This feedback is often expressed as a natural-language signal, sometimes viewed as a form of \emph{text gradient}: instead of providing numerical gradients over parameters, the environment provides textual feedback that suggests how the agent should revise its reasoning, strategy, or memory for future tasks. Existing sequential memory methods typically follow a retrieve--solve--update loop: the agent retrieves memory, solves the current task, and then uses the current trajectory and feedback to update the memory~\citep{suzgun2026dynamic,zhao2024expel,wei2025evo}. 
This design is intuitive and efficient, but it also introduces a fundamental risk. Each update is usually optimized locally for the most recent task, while its effect on the final memory state is rarely evaluated globally (Figure~\ref{fig:teaser}). As a result, a memory update that appears useful for the current task may overwrite previously useful knowledge, introduce noisy task-specific rules, or bias the memory toward recent examples. The key challenge is therefore not only how to generate memory updates, but also how to decide whether a proposed update should actually be deployed.


This creates a memory validation problem. Ideally, before deploying a candidate memory, the agent should estimate whether it improves performance beyond the current task. The most direct way to obtain such a signal is to compare the previous memory and the candidate memory on tasks that approximate the future task distribution. Since the future is unavailable, previously encountered tasks provide a natural proxy. However, validating every candidate memory on the full history is computationally impractical: the replay cost grows with the number of seen tasks and would introduce substantial latency after each update. A cheaper alternative is to compare memories at fixed intervals, such as every $N$ steps, but such schedules are heuristic and may miss harmful updates that occur between scheduled checks. Similarly, using only a small fixed replay set reduces cost but risks overfitting the deployment decision to stale examples. Thus, an effective memory controller must answer two coupled questions: \emph{when} should old and new memory be compared, and \emph{what} tasks should be used for comparison?

To solve the above challenges, we propose \textbf{Janus}, a plug-in memory controller that wraps existing sequential memory updaters and decides whether each candidate memory update should be accepted or rejected. Instead of treating every generated memory update as automatically beneficial, Janus views memory updating as a deployment decision: a candidate memory should replace the previous memory only when it is likely to improve the utility of the final memory for future tasks. Janus introduces two key designs to make this decision efficient. First, it uses a \textit{Memory Momentum Trigger} (MMT) to decide when old and new memory should be explicitly compared. Rather than triggering comparison after every task or at a fixed interval, MMT tracks the trajectory of memory changes and triggers comparison when a candidate update deviates substantially from the recent update direction, indicating that the update may either introduce useful new knowledge or distort the memory toward recent task-specific information. Second, when comparison is triggered, Janus evaluates the previous and candidate memories on a compact hybrid evaluation batch instead of replaying the full task history. This batch combines a stored support set, which includes both coverage tasks representing the seen task distribution and boundary tasks where memory choices previously changed correctness, with a fresh slice of recently encountered tasks to avoid overfitting the decision to a fixed support set. In this way, Janus improves memory deployment by selectively testing suspicious updates while keeping replay cost controlled.
Our main contributions are summarized as follows:
\vspace{-0.1in}
\begin{itemize}[leftmargin=0.35cm]
    \item \underline{\textbf{\textit{Memory deployment challenge.}}} We identify a key limitation of sequential LLM memory systems: locally generated memory updates are not necessarily globally useful, and blindly accepting them may produce final memories that are biased toward noisy task-specific information. \vspace{-0.1in}
    \item \underline{\textbf{\textit{Efficient plug-in memory control.}}} We propose Janus, a method-agnostic controller that wraps existing memory updaters and efficiently decides whether to accept or reject candidate memory updates. Janus combines a Memory Momentum Trigger with a compact hybrid evaluation batch over coverage, boundary, and fresh tasks, enabling old-versus-new memory selection without modifying the underlying updater or replaying the full task history. \vspace{-0.1in}
    \item \underline{\textbf{\textit{Strong empirical results.}}} Across six datasets, two backbone LLMs, and two sequential memory updater backbones, Janus consistently improves final memory usefulness, with average gains ranging from $+2.7$ to $+4.6$ percentage points over the corresponding base updaters.
\end{itemize}

\section{Method}

\subsection{Problem Setting}

\begin{figure*}[t]
    \centering
    \includegraphics[width = \textwidth]{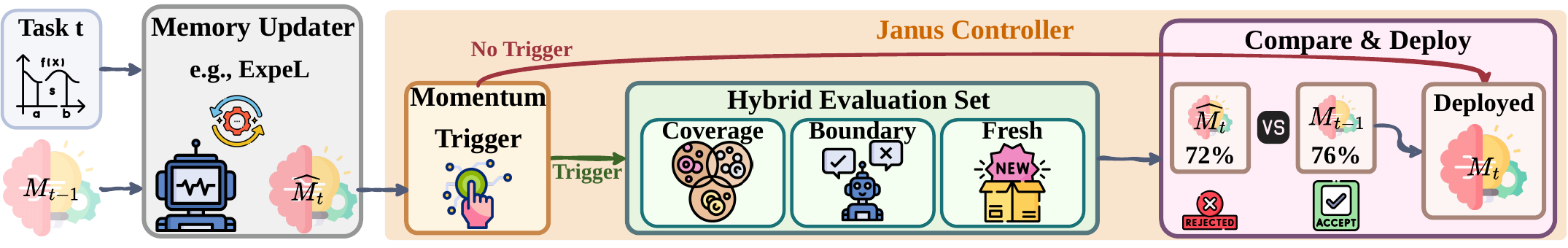}
    \caption{
    \textbf{Overview of Janus.}
    Given a task and the current memory $M_{t-1}$, a base updater proposes a candidate memory $\widehat{M}_t$. Janus acts as a plug-in controller that decides whether to deploy this candidate or retain the previous memory. It first uses a Memory Momentum Trigger to detect whether the candidate update deviates from the recent memory-update trajectory. If triggered, Janus compares $M_{t-1}$ and $\widehat{M}_t$ on a compact hybrid evaluation set composed of coverage, boundary, and fresh tasks, and deploys the memory with better evaluation performance. If not triggered, Janus accepts the candidate directly to avoid unnecessary replay.
    }
\vspace{-0.1in}
    \label{fig:janus}
\end{figure*}

Following~\citet{wei2025evo}, we consider a sequentially evolving memory setting where an LLM $\gL$ solves a stream of tasks while maintaining an external memory.
Let $\mathcal{D}=\{(x_t,y_t)\}_{t=1}^{T}$ denote the task sequence.
At step $t$, the model uses the previous memory $M_{t-1}$ to predict \(\hat{y}_t = \gL(x_t, M_{t-1}).\)
After receiving feedback $f_t$, such as correctness signals or textual critiques, a base memory updater proposes a revised memory:
\begin{equation}
    \widehat{M}_t =
    \texttt{Update}
    \left(
        M_{t-1},
        (x_t,\hat{y}_t,f_t),
        \gL
    \right).
\end{equation}
Existing sequential memory methods typically deploy this candidate directly by setting $M_t=\widehat{M}_t$.
In contrast, Janus treats memory updating as a deployment decision: given $M_{t-1}$ and $\widehat{M}_t$, it chooses
\begin{equation}
    M_t \in \{M_{t-1}, \widehat{M}_t\},
\end{equation}
with the goal of maintaining a final memory $M_T$ that generalizes well to future unseen tasks.

\subsection{Janus: Plug-in Memory Control}
As illustrated in Figure~\ref{fig:teaser}, blindly accepting every memory update can leave an LLM with a final memory that fails to support future unseen tasks. Motivated by this observation, we propose \textbf{Janus}, a plug-in controller that wraps a base memory updater and decides whether each candidate memory should be deployed. After solving the $t$-th task, the base updater proposes a candidate memory $\widehat{M}_t$ based on the previous memory $M_{t-1}$ and the current interaction. Rather than directly setting $M_t=\widehat{M}_t$, Janus chooses whether to accept the candidate memory or retain the previous memory.
This design addresses two core challenges in sequential memory control: \emph{when} to compare the previous and candidate memories, and \emph{what} tasks to use for comparison. For the first challenge, Janus introduces a Memory Momentum Trigger, which detects suspicious deviations in the memory-update trajectory and avoids unnecessary comparisons. For the second challenge, Janus uses a compact hybrid evaluation set, combining representative, memory-sensitive, and fresh tasks to approximate full-history replay with much lower cost.

\paragraph{Memory Momentum Trigger (MMT).}
Ideally, we would verify every memory update before deployment. However, comparing $M_{t-1}$ and $\widehat{M}_t$ after every task incurs substantial replay cost, while triggering comparisons at fixed intervals, e.g., every $N$ steps, is heuristic and may miss abrupt harmful updates. Janus instead views memory updates as a trajectory and triggers comparison only when the candidate update deviates substantially from recent memory evolution.
Let $\phi(\cdot)$ denote a text encoder that maps a memory state into a vector space. We represent the candidate update direction as
\begin{equation}
    z_t = \phi(\widehat{M}_t) - \phi(M_{t-1}).
\end{equation}
Janus maintains an exponential moving average of previous update directions:
\begin{equation}
    m_t = \beta m_{t-1} + (1-\beta)z_t,
\end{equation}
where $\beta$ controls the strength of memory momentum.
%
The intuition is that comparison is most valuable when a candidate update substantially changes the recent trajectory of memory evolution. Even when tasks arrive in a shuffled order, consecutive candidate updates may still induce aligned changes to the memory, corresponding to incremental refinements rather than major memory transitions. In such cases, repeatedly comparing old and new memories provides limited additional benefit. In contrast, a sharp directional deviation indicates that the candidate memory may significantly alter the deployed memory: it may introduce useful new knowledge, but it may also overwrite broadly useful information with recent-task-specific content. Janus therefore treats the historical momentum $m_{t-1}$ as a compact summary of recent memory evolution and uses directional misalignment as a signal for when explicit validation is needed.

Specifically, Janus triggers an old-versus-new memory comparison when
\begin{equation}
    \cos(z_t, m_{t-1}) < \tau,
\end{equation}
where $\tau$ is a threshold. If the trigger does not fire, Janus directly accepts the candidate memory, i.e., $M_t=\widehat{M}_t$. If the trigger fires, Janus evaluates $M_{t-1}$ and $\widehat{M}_t$ on a compact evaluation set and deploys the memory state with better performance.

\paragraph{Hybrid Trigger-Time Evaluation Set.}
When the memory momentum trigger fires, Janus needs to decide whether the candidate memory $\widehat{M}_t$ should replace the current memory $M_{t-1}$. A reliable comparison should approximate full-history replay, but evaluating on all previously seen tasks would make the replay cost grow linearly with the task stream. Janus therefore constructs a compact hybrid evaluation set that serves three goals: covering the global support of previously seen tasks, focusing on tasks that are sensitive to memory changes, and incorporating newly encountered tasks.
Formally, at trigger time $t$, Janus evaluates the two memory states on
\begin{equation}
    \mathcal{E}_t
    =
    \mathcal{S}^{\mathrm{cov}}_t
    \cup
    \mathcal{S}^{\mathrm{bdry}}_t
    \cup
    \mathcal{F}_t,
\end{equation}
where $\mathcal{S}^{\mathrm{cov}}_t$ is a coverage set, $\mathcal{S}^{\mathrm{bdry}}_t$ is a boundary set, and $\mathcal{F}_t$ is a fresh set. The coverage set summarizes the broad distribution of previously seen tasks, the boundary set stores memory-sensitive tasks where memory choices have previously changed the agent's behavior, and the fresh set injects tasks encountered since the last trigger. Together, these subsets provide a bounded-cost approximation to full-history replay.
Given $\mathcal{E}_t$, Janus evaluates both memories using the same LLM and deploys the memory with higher evaluation performance:
\begin{equation*}
    M_t
    =
    \argmax_{M \in \{M_{t-1}, \widehat{M}_t\}}
    \mathrm{Performance}(M; \mathcal{E}_t, \gL).
\end{equation*}
As the task stream evolves, Janus refreshes each subset of $\mathcal{E}_t$ to maintain an informative estimate of memory utility.
We update each subset as follows.

\noindent\textbf{Coverage set update.}
The coverage set $\mathcal{S}^{\mathrm{cov}}_t$ preserves representative support of the observed tasks. At each trigger time $t$, Janus refreshes this set by clustering the embeddings of all tasks seen so far:
\begin{equation}
    \mathcal{U}^{\mathrm{cov}}_t
    =
    \{x_1,\ldots,x_t\}.
\end{equation}
Using the previous centroids as initialization when available, Janus clusters $\{\phi(x):x\in\mathcal{U}^{\mathrm{cov}}_t\}$ and selects the task closest to each centroid as the new representative. This global refresh keeps the coverage set compact while ensuring that its representatives remain aligned with the full observed tasks.

\noindent\textbf{Boundary set update.}
The boundary set $\mathcal{S}^{\mathrm{bdry}}_t$ focuses the evaluation set on memory-sensitive tasks. After comparing $M_{t-1}$ and $\widehat{M}_t$ on $\mathcal{E}_t$, Janus identifies the flip set:
\begin{equation}
    \mathcal{B}_t
    =
    \left\{
        x \in \mathcal{E}_t :
        \mathbf{1}_{\mathrm{corr}}(x; M_{t-1})
        \neq
        \mathbf{1}_{\mathrm{corr}}(x; \widehat{M}_t)
    \right\},
\end{equation}
where $\mathbf{1}_{\mathrm{corr}}(x;M)\in\{0,1\}$ indicates whether the LLM $\gL$ answers task $x$ correctly when using memory $M$.
These tasks are informative because the choice of memory state directly changes the agent's correctness. Janus first removes tasks already selected into the coverage set:
\begin{equation}
    \widetilde{\mathcal{B}}_t
    =
    \mathcal{B}_t \setminus \mathcal{S}^{\mathrm{cov}}_t.
\end{equation}
It then fills the boundary set with tasks from $\widetilde{\mathcal{B}}_t$. If the number of new flip tasks is insufficient, Janus retains non-overlapping tasks from the previous boundary set:
\begin{equation}
    \mathcal{S}^{\mathrm{bdry}}_t
    \subseteq
    \widetilde{\mathcal{B}}_t
    \cup
    \left(
        \mathcal{S}^{\mathrm{bdry}}_{t^-}
        \setminus
        \mathcal{S}^{\mathrm{cov}}_t
    \right).
\end{equation}
This update rule preserves hard, behavior-changing examples while allowing the boundary set to evolve as new memory-sensitive regions are discovered.

\noindent\textbf{Fresh set update.}
Let $\ell(t)$ denote the time step of the most recent trigger before $t$. At trigger time $t$, Janus constructs the fresh set $\mathcal{F}_t$ by sampling from tasks encountered after $\ell(t)$:
\begin{equation}
    \mathcal{F}_t
    \subseteq
    \{x_{\ell(t)+1}, \ldots, x_t\}.
\end{equation}
This set prevents memory comparison from becoming a closed loop over a fixed replay buffer. By including newly observed tasks, Janus can test whether the candidate memory helps in recently explored regions of the task stream and can also discover new behavior-changing cases.

\section{Experiment}

\subsection{Experimental Settings}\label{sec:experiments_setting}

\begin{table*}[t]
\caption{\textbf{Main results on six datasets with two LLMs.}Accuracy (\%) is reported for each dataset. Janus is applied as a plug-in controller to two base memory updaters. Shaded rows indicate Janus-enhanced variants.}
\centering
\small
\tabcolsep = 2pt
\begin{tabular}{l c c c c c c c }
\toprule[1.2pt]
   \textbf{Method} & \textbf{MATH500} & \textbf{GPQA} & \textbf{MMLU-Pro (Eng.)}& \textbf{MMLU-Pro (Phy.)}&  \textbf{APIBench} &  \textbf{HumanEval} &  \textbf{Avg.}\\ 
    \midrule
    \rowcolor{gray!20} \multicolumn{8}{c}{\textbf{\emph{Qwen3-8B}}}\\ 
    \textbf{Memory-free} &78.2  &57.4  &61.2  &84.0  &65.2  &90.3  &72.7 \\ 
    \textbf{ExpRAG}&82.0  &69.2  &66.2  &87.2  &72.0  &91.2  &78.0 \\ \hdashline
    \textbf{DC-RS}&81.4  &73.4  &64.4  &84.0  &80.8  &93.2  &79.5 \\ 
    \rowcolor{gray!10}\textbf{DC-RS + Janus}&83.6  &81.5  &68.0  &89.2  &82.8  &93.9  &\textbf{83.2} \\ \hdashline
    \textbf{ExpeL}&80.0  &72.9  &68.8  &90.4  &66.4  &91.2  &78.3 \\ 
    \rowcolor{gray!10}\textbf{ExpeL + Janus}&81.6  &78.5  &71.6  &92.8  &70.4  &93.9  &\textbf{81.5} \\  \midrule
    \rowcolor{gray!20} \multicolumn{8}{c}{\textbf{\emph{DeepSeek-V4-Flash}}}\\ 
    \textbf{Memory-free} &75.4  &71.2  &68.4  &78.2  &66.8  &85.6  &74.3 \\ 
    \textbf{ExpRAG}&78.0  &74.9  &70.0 &82.4  &71.6  &86.4  &77.2 \\ \hdashline
    \textbf{DC-RS}&79.0  &70.7  &66.4  &83.2 &74.8  &86.0  &76.7 \\ 
    \rowcolor{gray!10}\textbf{DC-RS + Janus}&86.2  &73.3  &74.4  &89.2  &76.8  &87.8  &\textbf{81.3} \\  \hdashline
    \textbf{ExpeL}&74.8  &71.8  &73.2  &85.2  &82.0  &90.3  &79.6 \\ 
    \rowcolor{gray!10}\textbf{ExpeL + Janus}&79.2  &76.4  &75.6  &88.4  &83.2  &91.2  &\textbf{82.3} \\ 

\bottomrule[1.2pt]
\end{tabular}
\label{tab:main}
\end{table*}

\noindent\textbf{Datasets and LLMs.}
We evaluate Janus on six datasets covering mathematical reasoning (MATH500~\cite{hendrycks2021measuring,math500}), scientific reasoning (GPQA Diamond~\cite{rein2023gpqa}), professional STEM reasoning (MMLU-Pro Engineering and Physics~\cite{wang2024mmlu}), code generation HumanEval~\cite{chen2021evaluating}), and tool/API use (APIBench-HF~\cite{patil2024gorilla}). For memory construction, we sample training examples from each dataset: 500 for MATH, non-overlapping GPQA Main examples for GPQA Diamond, 250 examples for each MMLU-Pro subset, 50 HumanEval problems, and 250 APIBench-HF examples. Evaluation is conducted on the corresponding held-out test sets, with another 250 examples used for testing on MMLU-Pro and APIBench-HF. We focus on APIBench-HF because other APIBench subsets are relatively small and nearly saturated in our setting. Full task prompts are provided in Appendix~\ref{appx:prompt}. Our main experiments use Qwen3-8B~\cite{yang2025qwen3} and DeepSeek-V4-Flash~\cite{deepseekai2026deepseekv4}. Unless otherwise specified, we set \texttt{max-new-tokens=8192} and \texttt{temperature=0.7}; we enable thinking mode for Qwen3-8B and use non-thinking mode for DeepSeek-V4-Flash for cost efficiency.

\noindent\textbf{Baselines.}
We compare Janus with a representative set of sequential memory baselines under the same task stream.
\textbf{Memory-free} solves each task using only the underlying LLM without persistent memory, serving as a reference for measuring the benefit of external memory.
\textbf{ExpRAG}~\citep{wei2025evo} is a lightweight experience-reuse baseline that retrieves the top-$k$ most similar past tasks according to embedding similarity.
\textbf{DC-RS}~\citep{suzgun2026dynamic} maintains a dynamic cheatsheet and uses retrieval-and-synthesis to organize past experience into structured memory.
\textbf{ExpeL}~\citep{zhao2024expel} derives reusable insights from successful and failed trajectories through reflection; we adapt it to the sequential memory setting while following its original multi-attempt strategy.
For all base memory methods, we follow the original papers' recommended hyperparameter settings.
For Janus, unless otherwise specified, we use support-set size $K=20$, where $K'=12$ tasks are allocated to the coverage set $\mathcal{S}^{\mathrm{cov}}_t$ and the remaining $K-K'=8$ tasks are allocated to the boundary set $\mathcal{S}^{\mathrm{bdry}}_t$.
We set the fresh-set size to $K_{\mathcal{F}}=5$, the MMT threshold to $\tau=0.0$, and the momentum coefficient to $\beta=0.9$.
We provide detailed algorithmic descriptions in Appendix~\ref{appx:algo} and study Janus's sensitivity to key hyperparameters in Section~\ref{sec:hyperparam}.

\subsection{Main Results}

Table~\ref{tab:main} reports the main results across six datasets and two LLMs. We highlight two key observations.
\noindent\textbf{Structured memory updates are not always better than raw experience reuse.}
Compared with the memory-free baseline, most memory-based methods improve performance, showing the general value of external memory in sequential task solving.
However, stronger memory abstraction does not guarantee better final performance. Although DC-RS and ExpeL extract higher-level cheatsheets or reusable insights from past trajectories, they do not consistently outperform the lightweight ExpRAG baseline, which simply retrieves similar past experiences. This suggests that extracted memory can sometimes lose useful task-specific details, introduce noisy or over-specialized rules, or bias the final memory toward recent tasks. This observation motivates our central claim: the key issue is not only how to generate memory, but also whether each generated memory should be deployed.
\noindent\textbf{Janus consistently improves final memory usefulness across base updaters and LLMs.}
When applied to DC-RS and ExpeL, Janus improves the corresponding base updater across both LLMs and all datasets. On average, DC-RS+Janus improves over DC-RS from $79.5$ to $83.2$ on Qwen3-8B and from $76.7$ to $81.3$ on DeepSeek-V4. Similarly, ExpeL+Janus improves over ExpeL from $78.3$ to $81.5$ on Qwen3-8B and from $79.6$ to $82.3$ on DeepSeek-V4. The best-performing method in each LLM block is also a Janus-enhanced variant. These results support Janus as a method-agnostic controller: instead of replacing the base updater, it improves sequential memory systems by filtering locally generated updates and maintaining a more useful final memory for future tasks.


\subsection{MMT Trigger Ablation}

\begin{table}[t]
\centering
\small
\tabcolsep=2pt
\caption{
\textbf{Ablation study of the Memory Momentum Trigger.}
We compare Janus with alternative trigger policies using Qwen3-8B on GPQA and HumanEval. 
``Base'' denotes the base updater without Janus.
``Always'' compares old and new memory after every task, while ``Random'' and ``Periodic'' trigger comparisons under comparable trigger budgets to Janus.
}
\label{tab:mmt_trigger_ablation}
\begin{tabular}{l l cc cc}
\toprule
\multirow{2}{*}{\textbf{Updater}} 
& \multirow{2}{*}{\textbf{Policy}} 
& \multicolumn{2}{c}{\textbf{GPQA}} 
& \multicolumn{2}{c}{\textbf{HumanEval}} \\
\cmidrule(lr){3-4} \cmidrule(lr){5-6}
& & \textbf{Acc.} & \textbf{Trig. Rate} & \textbf{Acc.} & \textbf{Trig. Rate} \\
\midrule

\multirow{5}{*}{\textbf{DC-RS}}
& \textbf{Base}        & 73.4 & 0.0\%   & 93.2 & 0.0\%   \\
& \textbf{Random}             & 77.4 & 72.8\%     & 93.2 & 12.0\%      \\
& \textbf{Periodic}           & 76.4 & 50.2\%      & 92.1 & 12.0\%      \\
& \textbf{Always}             &80.0 & 100.0\% & 93.9 & 100.0\% \\
& \textbf{Janus}        & 81.5 & 72.4\%      & 93.9 & 12.0\%      \\

\midrule

\multirow{5}{*}{\textbf{ExpeL}}
& \textbf{Base}        & 72.9 & 0.0\%   & 91.2 & 0.0\%   \\
& \textbf{Random}             & 75.9 & 75.7\%      & 91.2  & 24.0\%    \\
& \textbf{Periodic}           & 74.3 & 50.2\%      & 93.0  & 20.0\%    \\
& \textbf{Always}             & 79.0 & 100.0\% & 94.7 & 100.0\% \\
& \textbf{Janus}        & 78.5 & 71.5\%      & 93.9  & 20.0\%      \\

\bottomrule
\end{tabular}
\end{table}

Table~\ref{tab:mmt_trigger_ablation} studies whether the Memory Momentum Trigger identifies useful moments for old-versus-new memory comparison.
We compare Janus with four alternatives: \emph{Base}, which never triggers comparison and directly accepts each candidate update; \emph{Always}, which compares after every task; \emph{Random}, which triggers comparison at each step with the same trigger probability as Janus; and \emph{Periodic}, which triggers every $N$ steps, with $N$ chosen to approximately match Janus's trigger count for the same updater and dataset.
This ablation targets the ``when to compare'' challenge: an effective controller should trigger comparison at informative memory-transition points rather than relying on a hand-designed schedule.

The results show that trigger timing matters.
Compared with random and periodic triggering, Janus achieves stronger or comparable accuracy under similar trigger budgets, especially on GPQA.
This suggests that MMT is not merely reducing the number of comparisons, but is able to capture meaningful changes in the memory-update trajectory where deployment decisions are more likely to affect final performance.
At the same time, Janus approaches the performance of the always-trigger policy with substantially fewer comparisons.
On HumanEval, Janus uses only a small fraction of triggers for both DC-RS and ExpeL while remaining close to always triggering.
On GPQA, Janus even outperforms always triggering for DC-RS, indicating that more frequent comparison is not necessarily better when the evaluation set is compact and potentially noisy.
Overall, the results support MMT as an efficient trigger mechanism for deciding when memory comparison is needed.

\subsection{Support Set Composition Ablation}

\begin{figure}[t]
\centering
\includegraphics[width=\linewidth]{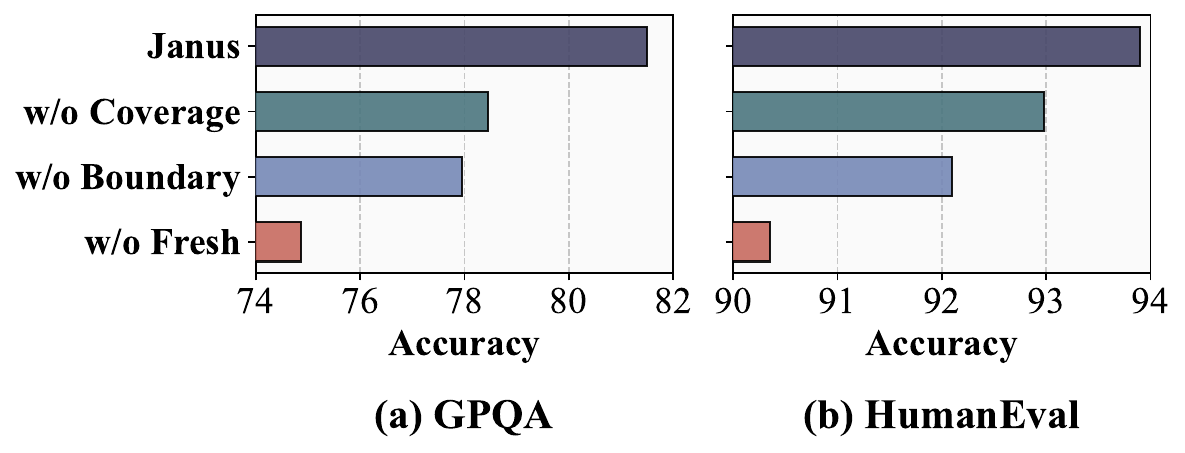}
\caption{
\textbf{Ablation study of the hybrid trigger-time evaluation set.}
We evaluate Janus with Qwen3-8B and the DC-RS updater on GPQA and HumanEval, reporting final test accuracy.
Each ablation removes one component: coverage, boundary, or fresh tasks.
}
\label{fig:support_set}
\vspace{-0.2in}
\end{figure}

Figure~\ref{fig:support_set} studies the composition of the hybrid evaluation set used when Janus triggers an old-versus-new memory comparison.
This ablation targets the ``what to compare'' challenge: after deciding that a candidate memory should be checked, Janus must evaluate it on a compact but informative set that approximates full-history replay.
We compare full Janus against three variants that remove coverage, boundary, or fresh tasks.
The results show that all three components contribute to reliable memory selection.
Removing the coverage set or the boundary set consistently reduces performance, suggesting that they provide complementary signals.
Coverage tasks preserve broad representativeness over the seen task distribution, while boundary tasks focus the comparison on memory-sensitive cases where different memory states are likely to change the prediction.
The largest drop comes from removing the fresh set, especially on both GPQA and HumanEval.
This indicates that relying only on the stored support set can make the comparison stale and self-reinforcing.
Fresh tasks provide recent pending evidence that has not yet been absorbed into the support set, allowing Janus to evaluate candidate memories on newly encountered regions of the task stream.
Overall, the ablation confirms that Janus's hybrid evaluation set is important for making effective memory deployment decisions under a bounded comparison budget.

\subsection{Memory Deployment Ablation}

\begin{figure}[t]
\centering
\includegraphics[width=\linewidth]{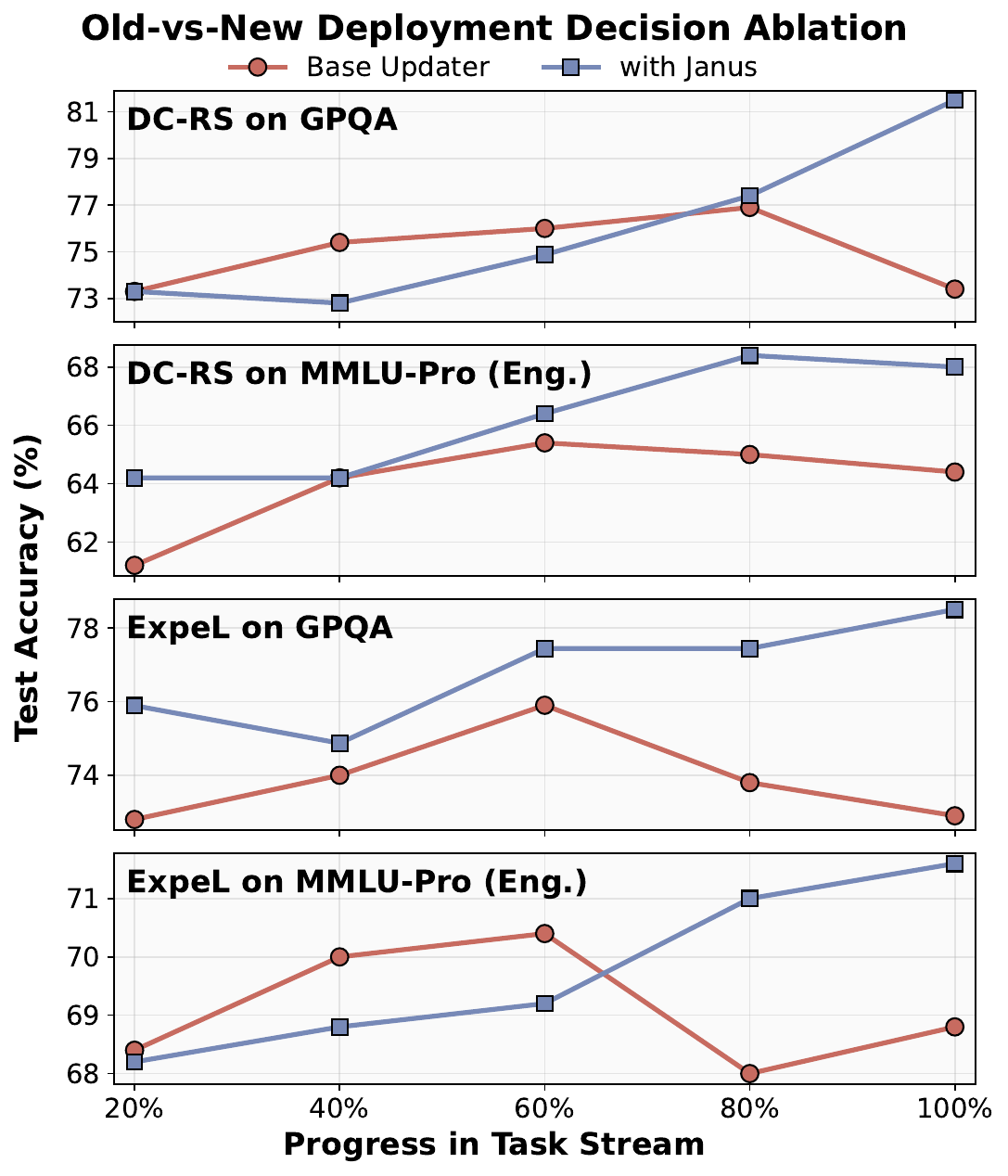}
\caption{
\textbf{Old-vs-new deployment decision ablation.}
We evaluate memory checkpoints obtained after processing $20\%$, $40\%$, $60\%$, $80\%$, and $100\%$ of the task stream using Qwen3-8B on GPQA and MMLU-Pro (Eng.).
The base updater directly deploys every candidate memory, while Janus selectively accepts or rejects candidate updates.
}
\label{fig:old_new_decision}
\vspace{-0.2in}
\end{figure}

Figure~\ref{fig:old_new_decision} evaluates whether the core deployment decision in Janus is useful. Instead of only reporting the final memory after the full stream, we measure the test accuracy of intermediate memory states after processing $20\%$, $40\%$, $60\%$, $80\%$, and $100\%$ of the training stream. This ablation directly tests whether selectively choosing between the previous memory and the candidate memory helps preserve useful information as more tasks are observed.

The results show that directly accepting every memory update does not necessarily lead to better future-task performance. For both DC-RS and ExpeL, the base updater often improves in the early or middle stages but then stagnates or drops as the stream continues. This indicates that additional memory updates can introduce noisy, overly specific, or conflicting information, causing the final memory to become less useful than earlier memory states. In contrast, Janus produces a more stable upward trend in most settings and achieves better final accuracy across all four comparisons. This supports the central motivation of Janus: sequential memory systems need a deployment controller, because the newest memory is not always the best memory for future tasks.
The improvement is especially clear near the end of the stream. While the base updaters can degrade after processing more tasks, Janus tends to preserve or improve test accuracy, suggesting that its old-vs-new decision helps retain broadly useful memory while filtering harmful candidate updates. Overall, this ablation confirms that Janus is not merely changing the update schedule; its explicit accept-or-reject decision is important for maintaining a final memory that generalizes better to unseen tasks. We further analyze robustness to different task-stream orders in Appendix~\ref{app:stream_order}, where Janus also yields more stable final-memory performance.

\subsection{Hyperparameter Sensitivity}
\label{sec:hyperparam}

\begin{table}[t]
\centering
\small
\setlength{\tabcolsep}{5pt}
\caption{
\textbf{Hyperparameter sensitivity of Janus-DC-RS.}
We sweep the support-set size $K=|\mathcal{S}_t|$ and the MMT threshold $\tau$ using Qwen3-8B on GPQA.
Unless otherwise specified, we use the default setting $K=20$, $K'=|\mathcal{S}^{\mathrm{cov}}_t|=12$, $|\mathcal{S}^{\mathrm{bdry}}_t|=K-K'=8$, $K_{\mathcal{F}}=|\mathcal{F}_t|=5$, and $\tau=0.0$.
Evaluation cost is approximated by $\#\mathrm{Trig.}\times(K+K_{\mathcal{F}})$.
}
\label{tab:hyperparam_sensitivity}
\begin{tabular}{@{}lcccccc@{}}
\toprule
\textbf{Sweep}
& $K$ 
& $\tau$ 
& \textbf{Acc.} 
& \textbf{\#Trig.} 
& \textbf{Trig. Rate} 
& \textbf{Cost} \\
\midrule

Default
& 20 
& 0.0
& 81.5
& 173 
& 72.4 
& 4,325 \\

\midrule

\multirow{3}{*}{$K$}
& 15 
& 0.0
& 74.9 
& 175 
& 74.8 
& 3,500 \\

& 25 
& 0.0 
& 76.9 
& 172 
& 76.8 
& 5,160 \\

& 50 
& 0.0
& 78.5 
& 157 
& 78.9 
& 8,635 \\

\midrule

\multirow{3}{*}{$\tau$}
& 20 
& -0.1
& 72.8 
& 152 
& 66.4 
& 3,800 \\

& 20 
& 0.1
& 77.4 
& 199 
& 86.9 
& 4,975 \\

& 20 
& 0.2
& 82.1
& 208 
& 90.8 
& 5,200 \\

\bottomrule
\end{tabular}
\end{table}

Table~\ref{tab:hyperparam_sensitivity} analyzes how Janus is affected by two key hyperparameters: the support-set size $K$ and the MMT threshold $\tau$.
These two parameters correspond to the two main efficiency dimensions of Janus.
The support-set size $K$ controls the coverage--boundary portion of the trigger-time evaluation set, while $\tau$ controls how frequently MMT triggers such comparisons.
We report final test accuracy together with the number of triggers and the estimated evaluation cost.
The results show that increasing the comparison budget does not always improve performance.
For the support-set sweep, $K=20$ gives the best accuracy among the tested values, while larger support sets increase evaluation cost without yielding better performance.
This suggests that a compact but informative support set is preferable to simply using more replay tasks.
For the trigger threshold sweep, increasing $\tau$ makes Janus more sensitive to memory-trajectory deviations and therefore increases the trigger rate.
A larger $\tau$ can improve accuracy, but it also incurs higher evaluation cost.
Overall, these results show that Janus provides a controllable trade-off between final memory performance and replay cost, with the default setting offering a strong balance between accuracy and efficiency.

\section{Related Work}
\noindent\textbf{Sequentially Evolving LLM Memory.}
Memory has become a central mechanism for enabling LLM agents to adapt over sequential interactions~\citep{xiang2026systematic,fang2025comprehensive,wei2025evo,chhikara2025mem0}. 
Recent memory-augmented agents extract reusable information from prior trajectories, feedback, reflections, workflows, or task solutions, and use such information to improve future reasoning and decision-making~\citep{wang2025far,fang2025memp,zhong2024memorybank,xu2025mem}. 
Representative approaches include reflection- or experience-based methods such as ExpeL~\citep{zhao2024expel}, dynamic memory construction methods such as Dynamic Cheatsheet~\citep{suzgun2026dynamic}, workflow-level memory methods such as Agent Workflow Memory~\citep{wang2025agent}, and retrieval- or structure-based memory systems such as G-Memory~\citep{zhang2026g} and Memento~\citep{zhou2025memento}. 
These methods mainly focus on how to construct or update memory, and assume that each newly generated memory should be directly deployed for future tasks. 
In contrast, Janus studies the deployment decision itself: given a candidate memory produced by an existing updater, should the agent accept it or keep the previous memory? 
This distinction is important because locally useful updates can still bias the final memory toward recent tasks or noisy task-specific rules, reducing its usefulness for future unseen tasks.

\noindent\textbf{Test-Time Adaptation and Memory Control.}
Test-time adaptation aims to improve model behavior during deployment, often by using feedback from the test environment rather than updating the model solely during offline training~\citep{wang2021tent,zhang2022memo}. 
Recent LLM agent systems extend this idea through natural-language feedback loops, where agents reflect on failures, revise solutions, refine prompts, or update strategies across interactions~\citep{shinn2023reflexion,madaan2023self,gou2024critic,asai2024self}. A related line of work formalizes such feedback as textual gradients, where natural-language critiques play a role analogous to gradients by indicating how an output, prompt, or policy should be mproved~\citep{yuksekgonul2025optimizing}. These methods show that feedback can support continual improvement without directly modifying model parameters.
However, feedback-driven updates are often local to the current task or recent trajectory, and may not improve the final memory used for future unseen tasks. Janus addresses this deployment challenge by selectively validating candidate updates with a momentum-based trigger and a compact hybrid evaluation set, deciding which feedback-induced memories should actually be deployed.

\section{Conclusion}
Sequentially evolving LLM memory raises a different scaling question from simply storing more experience or updating memory more often. As agents accumulate longer task histories, the critical issue becomes how to allocate additional inference effort: not every new memory should be trusted, and not every update deserves expensive validation. Janus points to a selective scaling strategy for self-evolving agents, where extra tokens are spent only when a candidate memory update is likely to change the trajectory of future behavior. By treating memory updating as a deployment decision, Janus shifts the focus from blindly accepting locally generated memories to controlling which memories should shape future inference. Across multiple datasets, backbone LLMs, and base memory updaters, Janus consistently improves final memory performance while avoiding unnecessary comparisons. Our ablations further show that both the trigger mechanism and the hybrid evaluation set are important for reliable memory deployment. This perspective suggests that robust sequentially evolving LLM memory requires not only better memory writers, but also mechanisms that decide when memory refinement is worth the cost.

\section*{Limitations}

\paragraph{Scope of memory mechanisms.}
Our evaluation focuses on prompt-based sequential memory systems, where the base LLM remains fixed and memory is updated through retrieval, summarization, reflection, cheatsheets, or other external text-based mechanisms. This scope matches many current memory-augmented agent systems and allows us to compare different updaters under a unified test-time protocol. However, another emerging direction is training-based evolution, where experience is used to update policies, skills, or model behavior through learning, such as reinforcement learning-based methods~\citep{yan2025memory,xia2026skillrl}. Extending Janus to jointly control external memory updates and parameter- or policy-level adaptation is an important direction for future work.

\paragraph{Evaluation coverage.}
Our experiments cover multiple task types and two backbone LLMs, but they do not exhaust all possible sequential agent environments. In particular, long-horizon interactive environments, multi-agent settings, and tasks with non-stationary distributions may introduce additional challenges for memory control. Further evaluation in these settings would help better characterize the generality of Janus.

\section*{Ethical considerations}

Our work focuses on improving the reliability of sequential LLM memory systems by controlling whether candidate memory updates should be deployed. In real-world applications, memory systems may store sensitive user information or incorrect task-specific rules, so they should be used with appropriate privacy protections, data filtering, and human oversight when necessary. We do not foresee major negative societal impacts from the proposed method itself, but responsible deployment depends on the underlying LLM, task domain, and memory contents


\bibliography{ref}

\clearpage
\appendix

\section{Stream Order Analysis} 
\label{app:stream_order}

\begin{figure}[t]
\centering
\includegraphics[width=\linewidth]{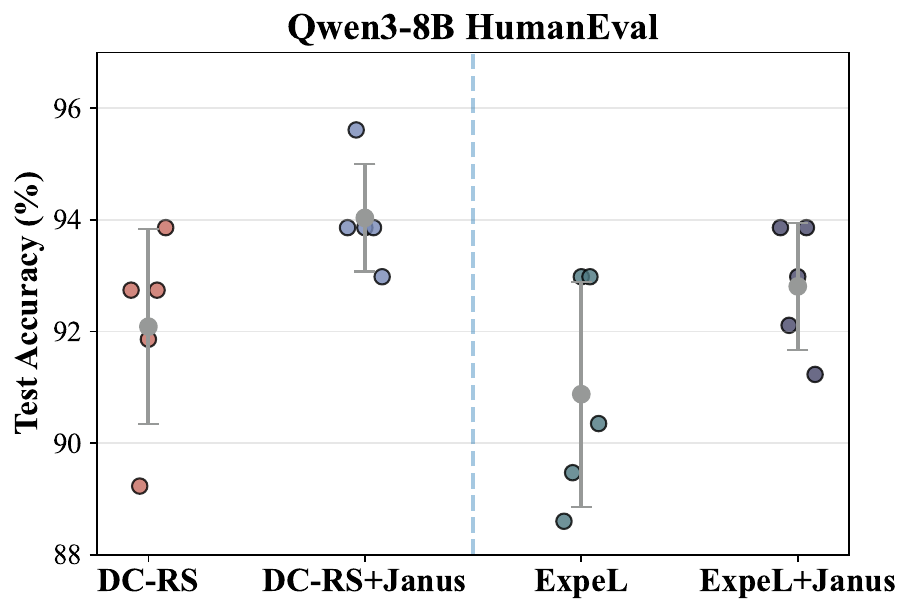}
\caption{
\textbf{Stream order analysis.}
We evaluate the final frozen-memory test accuracy under five different shuffled task-stream orders using Qwen3-8B on HumanEval.
Each point corresponds to one stream order, and error bars show the variation across orders.
}
\label{fig:order_sens}
\end{figure}

Figure~\ref{fig:order_sens} analyzes the sensitivity of sequential memory methods to the order of the task stream.
This experiment is closely related to our motivation: because base memory updaters revise memory using local feedback from the current task, the final memory may depend heavily on which tasks appear later in the stream.
We therefore shuffle the training stream with different random seeds and evaluate the final frozen memory on the same HumanEval test set.
The results show that Janus improves both accuracy and stability under stream-order changes.
For DC-RS, the base updater exhibits noticeable variance across different orders, indicating that directly accepting every candidate memory can lead to order-dependent final memories.
With Janus, the final performance is consistently higher and the variation across orders becomes smaller.
A similar pattern appears for ExpeL: the base updater is sensitive to the stream order, while ExpeL+Janus achieves stronger and more stable performance.

These results support the role of Janus as a memory-deployment controller.
By selectively accepting or rejecting candidate updates, Janus reduces the effect of myopic and recent-biased memory revisions.
Rather than letting the final memory be determined by the particular order in which tasks are encountered, Janus helps maintain a memory state that is more robust for future inference.

\section{Qualitative Case Study}
\label{app:qualitative_case}

We provide examples to illustrate how Janus behaves as a memory-deployment controller.
Each example is taken from a triggered decision during training.
When the Memory Momentum Trigger fires, Janus compares the previous memory $M_{t-1}$ and the candidate memory $\widehat{M}_t$ on the compact evaluation set, and then either rolls back to $M_{t-1}$ or deploys $\widehat{M}_t$.
For consistency, we use APIBench-HF for the DC-RS cases and MATH for the ExpeL cases.
For rollback cases, we show the candidate memory rejected by Janus; for accept cases, we show the previous memory that Janus replaces.

\begin{table}[t]
\centering
\small
\tabcolsep=3pt
\caption{
\textbf{Representative Janus memory-deployment decisions.}
All cases are triggered by MMT. Janus compares the previous memory and the candidate memory on the compact evaluation set and deploys the memory with better support-set performance.
}
\label{tab:qualitative_case}
\begin{tabular}{l l c c c}
\toprule
\textbf{Updater} & \textbf{Task} & \textbf{Old} & \textbf{New} & \textbf{Decision} \\
\midrule
DC-RS & APIBench-HF & 58.3 & 25.0 & Rollback \\
DC-RS & APIBench-HF & 53.3 & 86.7 & Accept \\
ExpeL & MATH & 73.3 & 53.3 & Rollback \\
ExpeL & MATH & 42.9 & 57.1 & Accept \\
\bottomrule
\end{tabular}
\end{table}

\paragraph{DC-RS rollback: rejecting a narrow recent-task rewrite.}
In the first APIBench-HF case, the current task asks the agent to classify news into categories such as technology, sports, or politics.
Before this step, the deployed memory $M_{t-1}$ contains API-use patterns from a recent robotics-related trajectory, including:
\begin{quote}
\small
\emph{Use Hugging Face pipelines for robotics task automation;} \\
\emph{Load pretrained Decision Transformer models for motor control tasks;} \\
\emph{Implement zero-shot reinforcement learning with multimodal transformers.}
\end{quote}
This previous memory is not necessarily ideal for every future APIBench task.
However, the candidate memory $\widehat{M}_t$ proposed by DC-RS rewrites the cheatsheet almost entirely around the latest news-classification task:
\begin{quote}
\small
\emph{Use zero-shot classification for multi-domain text categorization;} \\
\emph{Optimize classification with cross-encoder models;} \\
\emph{Utilize transformer-based pipelines for real-time classification.}
\end{quote}
The first rule directly reflects the current label space, while the remaining rules mostly repeat zero-shot text-classification advice.
Thus, the candidate does not merely add a useful insight; it shifts the whole memory toward one recent task type.
On the compact evaluation set, this rewrite reduces the support-set score from $58.3$ to $25.0$.
Janus therefore rolls back to $M_{t-1}$.
This case illustrates that Janus does not require the previous memory to be semantically perfect; it only needs to detect that the candidate deployment would make the memory less useful under the support-set comparison.

\paragraph{DC-RS accept: deploying a useful new tool pattern.}
The second APIBench-HF case shows the opposite behavior.
Here, the current task requires classifying product images of electronic devices.
The previous memory is dominated by multilingual text-classification rules:
\begin{quote}
\small
\emph{Use the Hugging Face pipeline API for text summarization with minimal code;} \\
\emph{Implement zero-shot classification with XLM-RoBERTa models for multilingual text categorization;} \\
\emph{Apply zero-shot classification with pre-trained multilingual models for cross-lingual categorization.}
\end{quote}
The candidate memory adds a new vision-oriented API pattern:
\begin{quote}
\small
\emph{Use CLIP-based zero-shot image classification for electronic device categorization;} \\
\emph{Deploy Hugging Face pipelines for zero-shot image classification with custom device categories;} \\
\emph{Apply zero-shot image classification with CLIP models for cross-domain product categorization.}
\end{quote}
Unlike the previous rollback case, this update introduces a genuinely new and reusable capability that the old memory lacks.
The support-set score increases from $53.3$ to $86.7$, so Janus accepts $\widehat{M}_t$.
This case shows that Janus is not conservative by default; it deploys candidate memories when they improve broader support-set behavior.

\paragraph{ExpeL rollback: rejecting a correct but narrow task-derived rule.}
In the first MATH case, the current problem asks for the greatest whole number that must divide the product of any three consecutive positive integers.
The previous ExpeL memory contains a diverse set of reusable mathematical heuristics, including:
\begin{quote}
\small
\emph{Combine like terms in polynomial expressions by grouping and summing coefficients;} \\
\emph{Use Fermat's Little Theorem for modular arithmetic when dealing with exponents modulo prime numbers;} \\
\emph{Apply Simon's Favorite Factoring Trick to solve equations by adding constants to factor expressions.}
\end{quote}
The candidate memory keeps most previous rules but adds the following rule:
\begin{quote}
\small
\emph{The product of any three consecutive positive integers is always divisible by 6.}
\end{quote}
This rule is mathematically correct, but it is tightly tied to the current problem pattern and has limited coverage over the support tasks used in this comparison.
In other words, correctness on the latest task does not imply that deploying the updated memory improves broader memory utility.
The support-set comparison confirms this: the previous memory scores $73.3$, while the candidate memory drops to $53.3$.
Janus therefore rejects the candidate and keeps the earlier memory.
This example shows that Janus can filter even correct local updates when they reduce the usefulness of the deployed memory for the broader task stream.

\paragraph{ExpeL accept: replacing a low-yield rule with a reusable procedure.}
The second MATH case involves converting repeating decimals into a common fraction.
Before the update, the last rule in the deployed memory is a simple percentage heuristic:
\begin{quote}
\small
\emph{Calculate percentage by dividing the count of desired elements by total elements and multiplying by 100.}
\end{quote}
The candidate memory replaces this low-yield rule with a procedural rule for converting repeating decimals:
\begin{quote}
\small
\emph{Convert repeating decimals to fractions by setting $x$ equal to the decimal, multiplying by $10^n$ where $n$ is the number of repeating digits, subtracting the original equation to eliminate the repeating part, and solving for $x$ as a fraction.}
\end{quote}
Although the two rules do not address the same mathematical topic, this is exactly the deployment decision Janus is designed to make under a bounded memory budget: whether the candidate memory state is more useful than the previous one. The new rule encodes a general algebraic procedure for a recurring class of problems, whereas the replaced percentage rule is a simpler heuristic with less support-set utility in this context. The support-set score improves from $42.9$ to $57.1$, so Janus accepts the candidate memory.

\paragraph{Takeaway.}
These cases show that Janus does not judge whether a candidate memory is locally correct or semantically similar to the replaced rule. Instead, it evaluates whether the resulting memory state is more useful on a compact support set. Thus, Janus can reject correct but narrow updates when they reduce broader utility, and accept task-motivated updates when they introduce reusable procedures or capabilities.

\clearpage
\section{Task Prompts}
\label{appx:prompt}

\newtcolorbox{promptbox}[1][]{
  breakable,
  colback=lightgraybox,
  colframe=black!40,
  title=\textbf{Evaluation Rubric Prompt},
  fonttitle=\bfseries,
  fontupper=\ttfamily\footnotesize,
  left=1mm,right=1mm,top=1mm,bottom=1mm,
  #1
}
\begin{promptbox}[title=\textbf{MATH Prompt}, width=\textwidth]
Solve the following math problem.

Please do your derivation in LaTeX as much as possible. Keep the reasoning concise but complete.
At the end, put only the final answer in the LAST \texttt{\textbackslash boxed\{...\}}.
If the final answer is a fraction, use LaTeX fraction form like \texttt{\textbackslash frac\{a\}\{b\}} (not decimal approximation unless required).

Problem: \{question\}
\end{promptbox}

\begin{promptbox}[title=\textbf{GPQA Prompt}, width=\textwidth]
Answer the following multiple-choice question.

Question: \{question\}

Choices:
\{choices\}

Think carefully and then output only the final option letter (A/B/C/D).
\end{promptbox}

\begin{promptbox}[title=\textbf{MMLU-Pro Prompt}, width=\textwidth]
Answer the following multiple-choice question.

Question: \{question\}

Choices:
\{choices\}

Think carefully and then output only the final option letter (\{valid\_labels\}).
\end{promptbox}

\begin{promptbox}[title=\textbf{HumanEval Prompt}, width=\textwidth]
Complete the following Python function. Implement \texttt{\{entry\_point\}} so that it satisfies the docstring.

\verb|```| python

\{question\}

\verb|```|

Return the full implementation (including the function signature) inside a single \verb|```| python... \verb|```| code block. Do not add any example usage or extra prose after the code block.
\end{promptbox}

\begin{promptbox}[title=\textbf{APIBench Prompt}, width=\textwidth]
You are a helpful API writer who can write APIs based on requirements.

\{question\}

Write a Python program in 1 to 2 lines to call API in \{framework\}.

The answer should follow the format: 
\verb|<<<|domain\verb|>>>| \$DOMAIN, \verb|<<<|api\_call\verb|>>>|: \$API\_CALL, \verb|<<<|api\_provider\verb|>>>|: \$API\_PROVIDER, \verb|<<<|explanation\verb|>>>|: \$EXPLANATION, \verb|<<<|code\verb|>>>|: \$CODE. 
Here are the requirements:

\begin{enumerate}
    \item \texttt{\$DOMAIN} should be inferred from the task description.
    \item \texttt{\$API\_CALL} should have only one line of code that calls the API.
    \item \texttt{\$API\_PROVIDER} should be the programming framework used.
    \item \texttt{\$EXPLANATION} should be a step-by-step explanation.
    \item \texttt{\$CODE} is the Python code.
    \item Do not repeat the format in your answer.
\end{enumerate}
\end{promptbox}


\section{Algorithms}
\label{appx:algo}

\begin{algorithm*}[ht]
\caption{ExpeL}
\begin{algorithmic}[1]
\REQUIRE Sequence of tasks $\gT=\{T_i\}_{i=1}^{N}$, large language model $\gL$,
self-reflection model $\gL_{\textsc{reflect}}$,
insight prompt $P_{in}$, retrieval budget $K$, step size $L$,
max tries $Z$
\STATE \textbf{Initialize:} Experience pool $\gB \gets F_{\textsc{manual}}$ (seed demos if available),
recent success set $\gS \gets \emptyset$, insight set $\hat{\iota} \gets \emptyset$

\FOR{Task $T_i=(x_i, y_i) \in \gT$}
    \STATE Retrieve Top-$K$ similar success cases $F_{\textsc{sim}}$ from $\gB$ \algcomment{Search}
    \STATE $\nu \gets \texttt{""}$ 
    \STATE $\tau^{\textsc{succ}} \gets \emptyset$, $\tau^{\textsc{fail}} \gets \emptyset$

    \FOR{$z=1$ \TO $Z$}
        \STATE $\hat{y}_i^{(z)} \gets \gL(x_i, F_{\textsc{sim}}, \hat{\iota}, \nu)$ \algcomment{Synthesis}
        \STATE $success^{(z)} \gets \mathrm{Evaluator}(x_i, y_i, \hat{y}_i^{(z)})$

        \IF{$success^{(z)} = 1$}
            \STATE $\tau^{\textsc{succ}} \gets (x_i, \hat{y}_i^{(z)})$
            \STATE $\gB \gets \gB \cup \{\tau^{\textsc{succ}}\}$ \algcomment{Evolve}
            \STATE \textbf{break}
        \ELSE
            \STATE $\tau^{\textsc{fail}} \gets (x_i, \hat{y}_i^{(z)})$ 
            \STATE $\nu \gets \textsc{Concat}(\nu,\; \gL_{\textsc{reflect}}(\tau^{\textsc{fail}}))$ \algcomment{Reflect}
        \ENDIF
    \ENDFOR

    \IF{$\tau^{\textsc{succ}} \neq \emptyset$}
        \STATE $\gS \gets \gS \cup \{\tau^{\textsc{succ}}\}$
    \ENDIF

    \IF{$\tau^{\textsc{succ}} \neq \emptyset \ \AND\  \tau^{\textsc{fail}} \neq \emptyset$}
        \STATE $\hat{\iota} \leftarrow \gL(P_{in}, \tau^{\textsc{succ}}, \tau^{\textsc{fail}}, \hat{\iota})$ \algcomment{Pair Update}
    \ENDIF

    \IF{$|\gS| = L$}
        \STATE $\hat{\iota} \leftarrow \gL(P_{in}, \gS, \hat{\iota})$ \algcomment{Batch Update}
        \STATE $\gS \gets \emptyset$
    \ENDIF
\ENDFOR
\end{algorithmic}
\end{algorithm*}

\begin{algorithm*}[ht]
\caption{Dynamic Cheatsheet Retrieval-and-Synthesis (DC-RS)}\label{alg:DC} %
\begin{algorithmic}[1]
\REQUIRE Sequence of tasks $\gT=\{T_i\}_{i=1}^{N}$, large language model $\gL$, retrival budget $K$, generator template $P_{gen}$, curator template $P_{cur}$ 
\STATE \textbf{Initialize:} Cheatsheet $\gM_0 \gets \emptyset$, history $\gH_0 \gets \emptyset$
\FOR{Task $T_i=(x_i, y_i) \in \gT$}
    \IF{$|\gH_{i-1}|>0$}
            \STATE Retrieve Top-$K$ similar past cases $C_{retr}$ from $\gH_{i-1}$ \algcomment{Search}
        \ELSE
            \STATE $\mathcal{P} \gets \emptyset$
        \ENDIF
\STATE Update memory state $\gM_{i} \gets \gL(P_{cur},\; C_{retr},\; x_i,\; \gM_{i-1})$ \algcomment{Evolve}
\STATE Answer $\hat{y}_{i} \gets \gL(P_{gen},\; x_i,\; \gM_{i})$ \algcomment{Synthesis}
\STATE Update history $\gH_{i} \gets \gH_{i-1}\cup \{x_i,\hat{y}_{i}\}$ \algcomment{Evolve}
\ENDFOR
\end{algorithmic}
\end{algorithm*}

\end{document}

%% file: math_commands.tex
\usepackage{amsmath,amsfonts,bm}

\def\eqref#1{equation~\ref{#1}}

\def\1{\bm{1}}

\DeclareMathAlphabet{\mathsfit}{\encodingdefault}{\sfdefault}{m}{sl}
\SetMathAlphabet{\mathsfit}{bold}{\encodingdefault}{\sfdefault}{bx}{n}

\def\gB{{\mathcal{B}}}

\def\gH{{\mathcal{H}}}

\def\gL{{\mathcal{L}}}
\def\gM{{\mathcal{M}}}

\def\gS{{\mathcal{S}}}
\def\gT{{\mathcal{T}}}

\DeclareMathOperator*{\argmax}{arg\,max}